# Towards Machine Translation for the Kurdish Language

**Sina Ahmadi**
Insight Centre for Data Analytics
National University of Ireland Galway
ahmadi.sina@outlook.com

**Mariam Masoud**
Insight Centre for Data Analytics
National University of Ireland Galway
maraim.elbadri@gmail.com

## Abstract

Machine translation is the task of translating texts from one language to another using computers. It has been one of the major tasks in natural language processing and computational linguistics and has been motivating to facilitate human communication. Kurdish, an Indo-European language, has received little attention in this realm due to the language being less-resourced. Therefore, in this paper, we are addressing the main issues in creating a machine translation system for the Kurdish language, with a focus on the Sorani dialect. We describe the available scarce parallel data suitable for training a neural machine translation model for Sorani Kurdish-English translation. We also discuss some of the major challenges in Kurdish language translation and demonstrate how fundamental text processing tasks, such as tokenization, can improve translation performance.

## 1 Introduction

Since the early advances in the field of natural language processing (NLP), one major task that motivated researchers to use machines for natural languages has been Machine Translation (MT). Natural language is notoriously complex and irregular with a great variability in the type of morphemes and their meanings as well as their syntactic and semantic dependencies in the context. Therefore, manual translation has proved to be inviable for such a task.

Over half a century, MT techniques have been getting more efficient and less language-dependent. Dictionary-based and rule-based approaches, which are deemed traditional approaches in the field, carry out the task of translation using manually-defined rules and resources (Tripathi and Sarkhel, 2010). Later, statistical machine translation further paved the way for diminishing the role of a linguist in the loop and decrease language dependency (Koehn, 2009). With the current advances in machine learning, particularly recurrent neural networks, neural machine translation (NMT) has been successfully used with state-of-the-art results for many languages (Koehn, 2020).

That being said, MT is not similarly challenging for all languages due to language-specific features. For instance, the translation of a morphologically-rich language, such as Czech or German, represents further challenges in alignment in comparison to a less morphologically-rich language like English (Passban, 2017; Mi et al., 2020). Moreover, MT systems require reasonably large aligned datasets and performant language processing tools such as tokenizer, stemmer and morphological analyzer (Koehn and Knowles, 2017). Such resources and tools are not always available, particularly for less-resourced languages. Languages are classified as less-resourced where general-purpose grammars and raw internet-based corpora are the only existing resources, lacking manually crafted linguistic resources and large monolingual or parallel corpora. Except the richer-resourced languages, the majority of the human languages are considered less-resourced (Cieri et al., 2016). This is also the current status of the Kurdish language, an Indo-European language spoken by 20-30 million speakers (Ahmadi et al., 2019; Esmaili and Salavati, 2013).

In this paper, we discuss the major challenges in MT for Sorani Kurdish, including the lack of basic language processing tools such a tokenization. To further highlight the challenges, we report the performance of two NMT models in various experimental setups based on the tokenization methods and resources. Despite the scarcity of parallel corpora for Kurdish, there are a few parallel resources which can be used for the task, partic-ularly the Tanzil corpus (Tiedemann, 2012) which contains 92,354 parallel sentences, the TED corpus (Cettolo et al., 2012) and KurdNet–the Kurdish wordnet (Aliabadi et al., 2014).

## 2 Related Work

There have been very few previous studies that address the Kurdish language in the MT realm. One of the outstanding projects in creating a rule-based machine translation system for Kurmanji and Sorani is the Apertium project (Forcada et al., 2011). In this open-source project, various tools and resources are developed for the Kurdish language, including bilingual and morphological dictionaries, structural transfer rules and grammars. Another initial attempt to create a machine translation system for Kurdish is `inKurdish`[1] which uses dictionary-based methods for translation. Taher et al. (2017) report that this system fails to translate based on the length of the input sentences and the degree of idiomaticity. As the two major existing machine translation tools for Sorani Kurdish, Kaka-Khan (2018) states that although the rule-based method of Apertium performs significantly better, limitations of the lexicon and transfer rules lead to incorrect translations and therefore, generalization across domains becomes difficult.

Kurdish language translation has been also of interest to many humanitarian organizations due to the refugee crisis in the current years, Translators without Borders (TWB)[2] and Tarjimly[3], to mention but a few (Balkul, 2018). Some of these organizations use mobile applications to enable refugees to get in touch with translators for their translation needs such as appointments with authorities. In the case of TWB, a machine translation system is created based on the Apertium project. However, the performance of the tool is not reported.

More recently, there has been an increasing number of resources created for the Kurdish language, such as dictionaries (Ahmadi et al., 2019), domain-specific corpora (Abdulrahman et al., 2019), folkloric corpus (Ahmadi et al., 2020a) and KurdNet–the Kurdish WordNet (Aliabadi et al., 2014). However, parallel corpora are more scarcely available. Bianet (Ataman, 2018) is a parallel news corpus containing 6,486 English-Kurmanji Kurdish and 7,390 Turkish-Kurmanji Kurdish sentences. Opus[4] also contains parallel translations in Kurmanji and Sorani for the GNOME and Ubuntu localization files[5](Espla-Gomis et al., 2019). More importantly, the Tanzil corpus provides translation of Qoranic verses in Sorani Kurdish.

In 2016, the translation service of Google, i.e. Google Translate[6], added Kurmanji Kurdish to its list of languages[7]. Motivated to explore this field for Sorani Kurdish, in this paper, we focus on creating a neural machine translation system for Sorani.

## 3 Sorani Kurdish

Sorani Kurdish is one of main dialects of Kurdish along with Kurmanji Kurdish and Southern Kurdish (Edmonds, 2013). This dialect is mainly spoken by the Kurdish populations in the Kurdish regions of Iran and Iraq. Unlike Kurmanji dialect for which a Latin-based script is used, Sorani Kurdish is mostly written in the Arabic-based script of Kurdish with no universally accepted orthography upon which scholars agree and is use by the public (Abdulrahman et al., 2019).

Kurdish has a subject-object-verb word order with a system of tense-aspect-modality and person marking (Haig and Matras, 2002). Moreover, Sorani Kurdish is a split-ergative language where transitive verbs in the past tenses are marked with an agentive case different from the nominative case (Manzini et al., 2015). The agentive case in Kurmanji Kurdish is the oblique case while Sorani Kurdish only uses different pronominal enclitics for ergative-absolutive alignment (Esmaili and Salavati, 2013). For further clarification, a few examples in Sorani Kurdish are provided below. In Example 1 in the past tense, the pronominal enclitic *=man* (in red) is used as the agentive marker and the suffix *in* (in green) is used for patient marking. In contrast, in Example 3, the same patient marker *-in* (in green) is used with a present tense as the subject marker with a nominative-accusative alignment and the pronominal enclitic *=man* (in red) is used in *małman* ‘our house’.

[1] `https://inkurdish.com`
[2] `https://translatorswithoutborders.org`
[3] `https://www.tarjim.ly`
[4] `http://opus.nlpl.eu`
[5] `https://l10n.gnome.org`
[6] `https://translate.google.com`
[7] Shortly after our project in August 2020, the Microsoft Translation service added Sorani and Kurmanji as well. See `https://www.bing.com/translator`

(1) *gułekanman hênan*. /. گوڵەکانمان هێنان<br>*gułekanman hênan* .<br>*guł=ek-an=man hêna-in*<br>flower.DEF.PL.1PL *bring*.PST.TR.ERG.3SG<br>'we brought the flowers.'

(2) *hênamanin*. /. هێنامانن<br>hênamanin .<br>hêna=man-in<br>bring.PST.TR.ERG.1PL.3SG<br>'we brought them.'

(3) deçin**e** małman. /. دەچنە ماڵمان<br>deçin e małman.<br>de-çi-in=e mał=man.<br>go-PRS.PROG.3PL=to house.N.1PL.<br>'(they) go/are going to our house.'

(4) *eme gułêke.* /. ئەمە گوڵێکە<br>*eme gułêk e* .<br>*em=e guł=êk e* .<br>this.DEM flower.IND is.3SG.COP.PRES .<br>'this is a flower.'

On the placement of agent markers, unlike patient markers, i.e. *-in*, which always appear immediately after the verb, the agentive markers follow an erratic pattern where they tend to appear immediately after the first prefix in verb forms, e.g. Example 2, or they attach to the leftmost morpheme in verbal phrases, if present, as in Example 1 (Walther, 2012; W. Smith, 2014). Moreover, Sorani Kurdish morphology is known to be complex, particularly due to the variety of affixes, clitics and the pattern in which they appear within the word and the phrase (Ahmadi and Hassani, 2020). Moreover, the stringing property of the Arabic-based script along with the lack of a unified orthography creates further complexity in a way that many word forms are concatenated into a single one. This is particularly the case of copula when emerges as an enclitic, as shown in Example 4. This yields further complications in the alignment of Kurdish and other languages.

Figure 1 illustrates the alignment of the English sentence "this is a woman from Canada" with its Sorani Kurdish translation *eme jinêke le Kanadawe*. The alignment is carried out at various levels, namely word-level, token-level and morpheme-level. This demonstrates how the granularity of the alignment varies depending on the level, ranging from a coarse-grained alignment at word-level where "is a woman" is aligned with only one word *jinêke*, to a more fine-grained alignment at token-level. Ultimately, at morpheme-level, circumposition *le* ... *ewe* and indefinite article *-êk* are alignable with their English equivalents, 'from' and 'a', respectively. To facilitate the reading, this example is provided in the Latin-based script of Kurdish.

## 4 Data Description

Given the scarcity of parallel corpora for Kurdish, we used all the available parallel corpora, despite their limited topic coverage and size. In this section, we describe the data used for this study.

### 4.1 Tanzil Corpus

Tanzil is a collection of Quran translations compiled by the Tanzil project[8]. There is one translation in Sorani Kurdish which is aligned with 11 translations in English making a total number of 92,354 parallel sentences with 3.15M words in the Sorani Kurdish side and 2.36M words in the English side. The corpus is available in Translation Memory Exchange (TMX) format where aligned verses are provided.

In the Kurdish translation, in addition to the translation of the verses, the interpretation of what is meant by the verse from the interpreter's point of view is provided. In addition, the Kurdish translation contains further idiomatic translations. These add to the granularity of the translations in the Kurdish side, while the translations in English are more conservative of the literal meaning. The interpretational texts are mostly specified in parentheses which make it feasible to remove automatically.

Some of the verses contain disjoined letters known as *Muqatta'at*. Although the English translation provides only a phonetic transliteration of such disjoined letters, the Sorani Kurdish one comes with chunks of text explaining the interpretation of such verses. For instance, the verse "الٓمٓ" in Arabic, is translated as "ALIF LAM MIM" in English, while the Kurdish verse contains explanatory sentences. In the same vein, complimentary phrases such as "peace be upon him" are mentioned throughout the English text, mostly in parentheses, while in the Kurdish translation, they are only partially specified from the actual content.

[8] http://tanzil.net

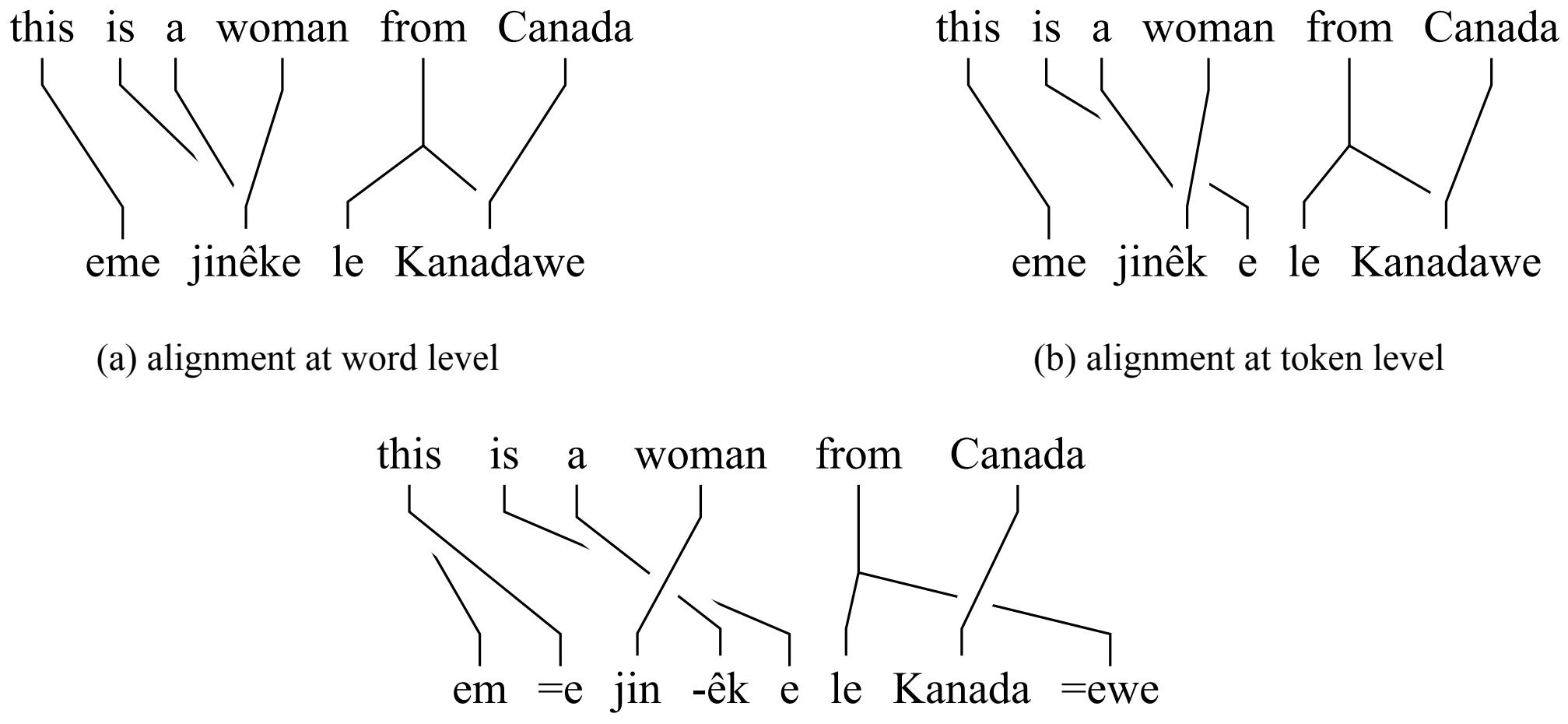

(a) alignment at word level

(b) alignment at token level

(c) alignment at morpheme level

Figure 1: Alignment of a Sorani Kurdish-English translation pair. The Latin-based script of Kurdish is used for facilitating the reading

One particular challenge regarding this corpus is the inconsistencies in writing Kurdish in the Arabic-based script. For instance, the conjunction ”و” (and) (written as ”û” in the Latin script), is frequently merged with the preceding word without considering the space between them. Moreover, the usage of punctuation marks is not thoroughly respected throughout the text. Due to the religious content of the data in Tanzil, there are many words which are written in Arabic in the Kurdish translation, particularly proper names like ”لوط” (Lot) which should be written as ”لوت” in the Kurdish script.

### 4.2 TED Corpus

The TED corpus[9] (Cettolo et al., 2012) is the collection of subtitles from TED Talks which are a series of high quality talks on ”Technology, Entertainment, and Design”. The Sorani Kurdish dialect is the only Kurdish dialect for which these subtitles are translated. Despite the small size of 2358 parallel sentences, the TED collection contains translations in a wider range of topics in comparison to Tanzil. Moreover, regarding the punctuation marks and orthography, this resource follows a more consistent approach. Although the sentences are aligned between Sorani and English, the alignments do not essentially correspond to a sentence. In some cases, a full paragraph containing many smaller phrases are aligned together. To further clarify this, Table 1 provides the average number of tokens per line where the average number of characters in the TED corpus is thrice the average of translations in Tanzil.

| Corpus | Language | tokens per line | characters per line |
|---|---|---|---|
| Tanzil | Kurdish | 25.82 | 159.36 |
| | English | 27.96 | 134.72 |
| Ted | Kurdish | 69.21 | 441.93 |
| | English | 93.54 | 452.88 |
| KurdNet | Kurdish | 7.51 | 44.27 |
| | English | 8.51 | 49.14 |

Table 1: Average number of tokens and characters per line in the English and Kurdish data

### 4.3 KurdNet–the Kurdish WordNet

WordNet (Miller, 1998) is a lexical-semantic resource which has been used in numerous natural language processing tasks such as word sense disambiguation and information extraction. In addition to semantic relationships such as synonymy, hyponymy, and meronymy, WordNet provides short definitions and usage examples for groups of synonyms, also known as synsets. KurdNet–the Kurdish WordNet (Aliabadi et al., 2014) is created based on a semi-automatic approach centred around building a Kurdish alignment for Base Concepts, which is a core subset of major meanings in WordNet. The current version of KurdNet contains 4,663 definitions which are directly translated from the Princeton WordNet (version 3.0). Although the number of the translated definitions is trivial for the task of machine translation, we included this resource as it contains more domain-specific terms, for instance in biology or philosophy, and it also reflects a more modern usage of the language in comparison to the religious content of Tanzil corpus.

[9] https://wit3.fbk.eu

# 5 Experiment Settings

## 5.1 Data Preparation

In order to remove non-relevant characters and clean the data, we unify the encoding of the characters by converting similar graphemes to unique ones, as described in (Ahmadi, 2019). The Arabic script is adapted to many languages, including Kurdish, where many graphemes may look alike but have different encoding. For instance, despite the similarity of ”ك” and ”ي” respectively to ”ک” and ”ی” , only the latter ones are used in Kurdish. Moreover, zero-width non-joiner character (`U+200C`) are removed and non-Kurdish characters used in proper names are replaced with the Kurdish equivalents, e.g. ”ط” with ”ت”. We also carried out an orthographic normalization throughout all the corpora by replacing initial ”ر” (r) with ”ڕ” (ř). Although the first one does not occur in Sorani Kurdish, some orthographies suggest using it and therefore, create variations in Sorani Kurdish texts. Moreover, interpretational texts provided between parentheses are removed.

On the English side of the data, the normalization step is consisted of removal of text within parentheses and truecasing. It should be noted that the Arabic script does not have character case.

## 5.2 Tokenization

The task of tokenization is of high importance in various tasks in NLP, particularly in machine translation (Domingo et al., 2018). In many languages including Kurdish, spaces are used to determine the boundary of tokens. However, due to the lack of a universal orthography and the complexity of Kurdish morphology, more than one token can sometimes be concatenated into one without any space.

At the time of carrying out this research, no tokenization tool was available for Kurdish language. Therefore, we trained three tokenization models using the state-of-the-art unsupervised tokenization methods provided by HuggingFace Tokenizers [10] and SentencePiece[11] (Wu et al., 2016). In the first case, we used `WordPiece` which is a subword tokenization algorithm used for BERT language model (Devlin et al., 2018). In the latter, we trained two models: byte-pair-encoding (BPE) and unigram language model (Unigram). All the models are trained with the `vocabulary_size=50000` and `character_coverage=1.0` using the available Sorani Kurdish raw corpora, namely PEWAN corpus containing 18M words (Esmaili et al., 2013), the Kurdish Textbooks Corpus (KTC) containing 693,800 words, (Abdulrahman et al., 2019), Veisi et al's corpus containing 8.1M words (Veisi et al., 2020) and Sorani Kurdish folkloric lyrics corpus containing 49,582 words (Ahmadi et al., 2020a). We preprocessed these corpora following the text normalization step described earlier. Additionally, we used a regular expression tokenisation method based on the WordPunct tokenizer of NLTK (Loper and Bird, 2002).

To remedy the systematic concatenation of conjunction ”و” (and) in the Tanzil corpus, we carried out an additional step where the frequency of the words ending with and without ”و” (û) is calculated in the PEWAN corpus (Esmaili et al., 2013). If the frequency of the word form without ”و” is higher than the word form with ”و”, we consider that the conjunction is meant and therefore we split the word into two tokens. For instance, ”تاوانبارو” is an incorrect word composed of two words ”تاوانبار” (guilty) and ”و” (and). In PEWAN, the original word has a frequency of 5 against 1218 for ”تاوانبار” in the same corpus. Therefore, applying this step yields a space between the two words and replaces them by the initial incorrect word. Figure A.3 in Appendix A provides normalized example pairs in English and Kurdish and their changes after this step.

## 5.3 Models

The experiment was performed using py-Torch version of OpenNMT (Klein et al., 2017), which is an open source library for training and deploying sequence to sequence NMT models. We deployed two variations of model settings: Model 1 and Model 2. The base model, Model 1, is set with the following hyper-parameters: two LSTM (Long Short Term Memory) layers with 200 hidden units for both the encoder and the decoder. The second model, Model 2, is the default OpenNMT model with two hidden LSTM (Long Short Term Memory) layers and 500 hidden units per layer on both the encoder and the decoder, batch size of 64, and 0.3 dropout probability and word embeddings of

[10] https://github.com/huggingface/tokenizers

[11] https://github.com/google/sentencepiece

| Tokenizer | Dataset | Number of sentences (tokens) | | | | |
|---|---|---|---|---|---|---|
| | | All | Train | Validation | Test 1 | Test 2 |
| BPE | Tanzil | 92325 (3335725) | 66476 (2406706) | 8308 (296874) | 8309 (297237) | 9232 (334908) |
| | TED | 2355 (253777) | 1697 (185258) | 212 (22324) | 211 (21879) | 235 (24316) |
| | KurdNet | 4659 (46384) | 3357 (33542) | 418 (4054) | 419 (4178) | 465 (4610) |
| | **All** | | 71532 (2625506) | 8940 (323252) | 8941 (323294) | 9932 (363834) |
| Unigram | Tanzil | 92325 (3365517) | 66476 (2428059) | 8308 (299634) | 8309 (299765) | 9232 (338059) |
| | TED | 2355 (260015) | 1697 (189879 | 212 (22860) | 211 (22377) | 235 (24899) |
| | KurdNet | 4659 (46336) | 3357 (33491) | 418 (4055) | 419 (4171) | 465 (4619) |
| | **All** | | 71532 (2651429) | 8940 (326549) | 8941 (326313) | 9932 (367577) |
| WordPiece | Tanzil | 92325 (3348264) | 66476 (2415538) | 8308 (298174) | 8309 (298321) | 9232 (336231) |
| | TED | 2355 (247865) | 1697 (180822) | 212 (21773) | 211 (21615) | 235 (23655) |
| | KurdNet | 4659 (46228) | 3357 (33391) | 418 (4063) | 419 (4168) | 465 (4606) |
| | **All** | | 71532 (2629751) | 8940 (324010) | 8941 (324104) | 9932 (363742) |
| WordPunct | Tanzil | 92325 (2909512) | 66476 (2098910) | 8308 (258926) | 8309 (259381) | 9232 (292295) |
| | TED | 2355 (250617) | 1697 (183596) | 212 (21886) | 211 (21353) | 235 (23782) |
| | KurdNet | 4659 (38950) | 3357 (28130) | 418 (3450) | 419 (3514) | 465 (3856) |
| | **All** | | 71532 (2310636) | 8940 (284262) | 8941 (284248) | 9932 (319933) |

Table 2: Number of sentences and tokens (in parentheses) of the Kurdish datasets based on various tokenization models and testing scenario

100 dimension. Regarding the word embeddings, we used the FastText pre-trained word vectors for Kurdish (Mikolov et al., 2018) and GloVe word embeddings trained on 6B tokens for English (Pennington et al., 2014).

### 5.4 Evaluation

The performance of the models is evaluated using the following three evaluation metrics; BLEU (Papineni et al., 2002), METEOR (Lavie and Agarwal, 2007) and TER (Snover et al., 2006). BLEU (Bilingual Evaluation Understudy) is an evaluation metric that matches $n$-grams from multiple metric for evaluation of translation with explicit ordering, and METEOR (Metric for Evaluation of Translation with Explicit ORdering) is based on the harmonic mean of precision and recall. TER (Translation Error Rate) is a metric that represents the cost of editing the output of the MT systems to match the reference. High score of BLEU and METEOR means the system produces a highly fluent translation, but a high score of TER is a sign of more post-editing effort and thus the lower the score the better.

## 6 Results and Analysis

To analyze the performance of the models and evaluate the impact of tokenization on the translation quality, we create various datasets based on the tokenization techniques, namely BPE, Unigram, WordPiece and WordPunct. As the Tanzil corpus is remarkably larger, we create two sets of testing scenarios where initially 10% of each dataset is set aside as the first testing set (Test 2 in Table 2). This way, the performance of the model with respect to each dataset can be evaluated separately as well. The remaining data are then merged all together and split into train, test and validation sets with 80%, 10% and 10% proportions respectively. The test set in the latter scenario is specified as Test 1 in Table 2[12].

### 6.1 Quantitative Analysis

Table 3 presents the performance of our two neural translation models, Model 1 and Model 2, with respect to the two test sets, Test 1 and Test 2, and various unsupervised neural tokenization models.

Regarding Test 1, in both Kurdish to English and English to Kurdish translations, the WordPunct tokenization model has the highest results in BLEU and METEOR and the lowest with respect to TER. Surprisingly, Model 2 which is trained with better hyper-parameters, performs better only in English to Kurdish translation while Model 1 provides the best results for Sorani and English translation.

Regarding Test 2 where 10% of each parallel corpus is used for testing purpose, our trained models perform relatively good with respect to the Tanzil corpus. However, all the setups fail to translate KurdNet and TED corpora efficiently, in such a way that the TER score is 1 in almost all cases. We believe that such a mediocre performance is due to (a) imbalance of the data, as most of the parallel sentences are provided from the Tanzil corpus, (b) type of sentences in KurdNet and the quality of alignments in TED (see Table 1), and (c) domain-specific terms which are used in the KurdNet and TED corpora while more generic words are used in the Tanzil corpus. In comparison to the Tanzil and TED corpora, WordNet definitions are significantly short. Moreover, synsets definitions are more objective and contain technical words. In other words, words which are more frequently used in subjective texts, such as pronouns, are less observed in this resource.

[12]All the datasets are available at `https://github.com/sinaahmadi/KurdishMT`

| Corpus | | Tokenization | Model 1 | | | | | | Model 2 | | | | | |
|---|---|---|---|---|---|---|---|---|---|---|---|---|---|---|
| | | | `ckb-en` | | | `en-ckb` | | | `ckb-en` | | | `en-ckb` | | |
| | | | BLEU | METEOR | TER | BLEU | METEOR | TER | BLEU | METEOR | TER | BLEU | METEOR | TER |
| Test 2 | Tanzil | WordPiece | 21.02 | 0.24 | 0.61 | 51.44 | 0.3635 | 0.32 | 19.48 | 0.231 | 0.64 | 50.48 | 0.3616 | 0.32 |
| | | Unigram | 20.71 | 0.2381 | 0.6 | 50.21 | 0.3582 | 0.32 | 19.53 | 0.232 | 0.63 | 50.95 | 0.3613 | 0.32 |
| | | WordPunct | **22.03** | **0.2454** | **0.58** | 58.36 | 0.412 | **0.27** | 20.42 | 0.2384 | 0.61 | **59.28** | **0.4156** | **0.27** |
| | | BPE | 21.03 | 0.2392 | 0.61 | 50.04 | 0.3588 | 0.32 | 19.49 | 0.2315 | 0.63 | 50.28 | 0.358 | 0.33 |
| | KurdNet | WordPiece | 5.86 | 0.1245 | 0.93 | **3.9** | 0.091 | **0.99** | **6.47** | **0.1297** | 0.9 | 3.25 | 0.0904 | 1.01 |
| | | Unigram | 5.88 | 0.1216 | 0.91 | 3.38 | 0.0884 | 1 | 6.15 | 0.1269 | **0.89** | 3.82 | **0.0923** | 1 |
| | | WordPunct | 5.81 | 0.1169 | 0.9 | 2.57 | 0.082 | 1 | 5.16 | 0.1242 | 0.9 | 2.82 | 0.0867 | 1 |
| | | BPE | 6.32 | 0.1209 | 0.92 | 3.5 | 0.0853 | 1 | 6.39 | 0.133 | 0.9 | 3.05 | 0.0826 | 0.99 |
| | TED | WordPiece | **1** | **0.0875** | 0.9 | 0 | 0.0378 | 0.99 | 0.74 | 0.0758 | 0.9 | **0.05** | 0.0383 | 0.99 |
| | | Unigram | 0.88 | 0.0775 | 0.91 | 0 | **0.0415** | **0.97** | 0.89 | 0.0863 | **0.89** | 0 | 0.0397 | 0.99 |
| | | WordPunct | 0.62 | 0.072 | **0.89** | 0 | 0.0295 | 1 | 0.59 | 0.0712 | 0.9 | 0 | 0.0289 | 0.99 |
| | | BPE | 0.92 | 0.0853 | 0.91 | 0 | 0.0353 | 0.99 | 0.75 | 0.0803 | 0.9 | 0 | 0.0298 | 0.99 |
| Test 1 | | WordPiece | 19.05 | 0.2242 | 0.65 | 46.47 | 0.3322 | 0.37 | 17.49 | 0.2153 | 0.67 | 45.23 | 0.328 | 0.38 |
| | | Unigram | 18.95 | 0.2235 | 0.63 | 45.24 | 0.3275 | 0.38 | 17.47 | 0.2156 | 0.66 | 45.83 | 0.3299 | 0.38 |
| | | WordPunct | **19.95** | **0.2276** | **0.61** | 52.21 | 0.3726 | **0.33** | 18.5 | 0.2222 | 0.65 | **52.94** | **0.3753** | **0.33** |
| | | BPE | 19.06 | 0.2233 | 0.63 | 45.13 | 0.3282 | 0.38 | 17.51 | 0.2157 | 0.66 | 45.14 | 0.3269 | 0.38 |

Table 3: Quantitative results for the evaluation of Kurdish $\leftrightarrow$ English using various test sets. Results in bold represent the best system within the given models

To further clarify the poor results of TED which particularly has a large number of tokens per sentence, we carried out a set of experiments by filtering sentences based on their number of tokens. For this purpose, we created four smaller datasets based on the TED test sets (Test 2) containing a maximum number of 25, 50, 75 and above 75 tokens and evaluated the performance of the models using various $n$-grams for the BLEU score. Figure 2 demonstrates how a lower number of tokens per sentence improves the BLEU score significantly. That said, the overall performance of the models is still not satisfying, with the best model

In all the testing scenarios, the English-Kurdish models significantly outperform the Kurdish-English translation. This is explainable as there is only one translation available for Kurdish in Tanzil but 11 translations for English. In other word, a sentence in Kurdish is aligned with 11 different sentences in English.

## 6.2 Qualitative Analysis

Figures A.4 and A.5 illustrate a few translations in our parallel corpora along with their back-translation. Despite the poor performance of the models with respect to TED and KurdNet, the system translations often carry meaning in a comprehensible way. In other words, even if the system translations do not correspond to the reference ones, they are not completely nonsensical, depending on the tokenization method.

Interestingly, some of the system translations are correct, even if the reference translations were not originally correctly-written. This is particularly the case of the Tanzil corpus. For instance, "you are a people unknown to me" in Figure A.4, is correctly translated in Kurdish while the Kurdish translation is written without any space in both the system output and the reference translation. In the same vein, we observe that the trained models capture information regarding synonyms or semantically-related words. For instance, 'knowledge' is translated as زانست (*zanist*) 'science' in a reference translation, while زانیاری (*zanyarî*) 'knowledge' is used for the same word by our models.

# 7 Conclusion and Future Work

In this paper, we present our efforts to develop a neural machine translation system for the Sorani dialect of Kurdish. We describe how due to scarcity of parallel corpora, we used translations of religious texts as the material for training a machine translation system. Moreover, we created basic language processing tools, such as tokeniza-tion, by using unsupervised techniques, namely WordPiece, byte-pair-encoding and unigram language model. We train two neural machine translation models using different hyper-parameters and evaluate the models based on the datasets and the tokenization techniques. Although the imbalanced data makes the models over-fit, our qualitative analysis indicates that some syntactic and lexical properties of Kurdish are correctly learnt in the translation outputs.

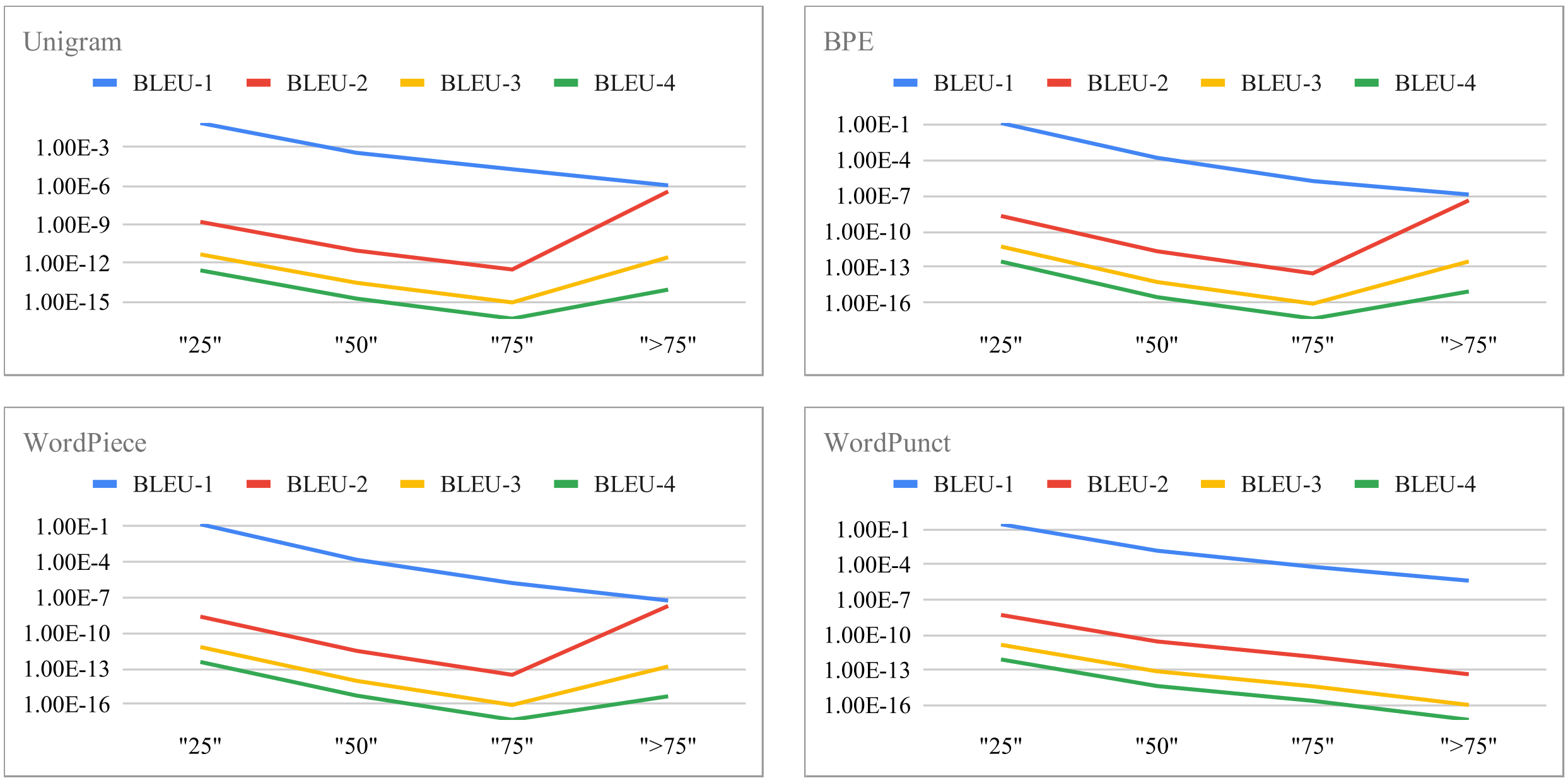

Figure 2: The performance of Model 1 in translating Kurdish to English with respect to certain length-limited sentences in the Kurdish TED corpus in terms of various BLEU scores

There are two major limitations in the current project which could not be addressed due to our focus being on the preprocessing steps: a baseline system and further experiments with respect to hyper-parameters. Given the current state of lack of parallel corpora, we were not able to extend the study the other dialects. We strongly believe that this should be a motivation to create more resources for the Kurdish language[13]. Moreover, as an initial work of its kind for the machine translation of Kurdish, we dealt with many basic language processing tasks which were not properly addressed. Developing such tools should be a priority in the field of Kurdish language processing.

Regarding future work, we would like to suggest morpheme-based translation (Luong et al., 2019). As Kurdish is a morphologically-rich language, it might be beneficial to go beyond tokens and carry out the alignment task at morpheme-level. We also believe that lexicons can be efficiently incorporated for compensating the scarcity of resources for Kurdish (Zhang and Zong, 2016). We also propose the usage of other dialects of Kurdish, such as Kurmanji, and closely-related languages, like Persian, for improving the performance of future machine translation models (Nakov and Tiedemann, 2012). Another recent promising direction for a low-resource setup like Kurdish is monolingual sequence-to-sequence pretraining techniques, such as MAsked Sequence to Sequence pre-training (MASS) (Song et al., 2019) or *mBART* (Liu et al., 2020).

[13]Shortly after this project, Ahmadi et al. (2020b) present a parallel corpus created based on multilingual news websites content.

## 8 Acknowledgements

The authors would like to thank Dr. Diego Moussallem for his collaboration at the initial stages of this work.

## References

Roshna Omer Abdulrahman, Hossein Hassani, and Sina Ahmadi. 2019. Developing a Fine-Grained Corpus for a Less-resourced Language: the case of Kurdish. *arXiv preprint arXiv:1909.11467*.

Sina Ahmadi. 2019. A rule-based Kurdish text transliteration system. *Asian and Low-Resource Language Information Processing (TALLIP)*, 18(2):18:1–18:8.

Sina Ahmadi and Hossein Hassani. 2020. Towards Finite-State Morphology of Kurdish. *arXiv preprint arXiv:2005.10652*.

Sina Ahmadi, Hossein Hassani, and Kamaladdin Abedi. 2020a. A Corpus of the Sorani Kurdish Folkloric Lyrics. In *Proceedings of the 1st Joint Spoken Language Technologies for Under-resourced languages (SLTU) and Collaboration and Computing for Under-Resourced Languages (CCURL) Workshop at the 12th International Conference on Language Resources and Evaluation (LREC)*, Marseille, France.

Sina Ahmadi, Hossein Hassani, and Daban Q Jaff. 2020b. Leveraging Multilingual News Websites for Building a Kurdish Parallel Corpus. *arXiv preprint arXiv:2010.01554*.

Sina Ahmadi, Hossein Hassani, and John P. McCrae. 2019. Towards electronic lexicography for the Kurdish language. In *Proceedings of the sixth biennial conference on electronic lexicography (eLex)*, pages 881–906, Sintra, Portugal.

Purya Aliabadi, Mohammad Sina Ahmadi, Shahin Salavati, and Kyumars Sheykh Esmaili. 2014. Towards building kurdnet, the kurdish wordnet. In *Proceedings of the Seventh Global Wordnet Conference*, pages 1–6.

Duygu Ataman. 2018. Bianet: A parallel news corpus in turkish, kurdish and english. *arXiv preprint arXiv:1805.05095*.

Halil İbrahim Balkul. 2018. A Comparative Analysis Of Translation/interpreting Tools Developed For Syrian Refugee Crisis. *International Journal of Language Academy*.

Mauro Cettolo, Christian Girardi, and Marcello Federico. 2012. Wit3: Web inventory of transcribed and translated talks. In *Conference of european association for machine translation*, pages 261–268.

Christopher Cieri, Mike Maxwell, Stephanie Strassel, and Jennifer Tracey. 2016. Selection criteria for low resource language programs. In *Proceedings of the Tenth International Conference on Language Resources and Evaluation (LREC'16)*, pages 4543–4549.

Jacob Devlin, Ming-Wei Chang, Kenton Lee, and Kristina Toutanova. 2018. Bert: Pre-training of deep bidirectional transformers for language understanding. *arXiv preprint arXiv:1810.04805*.

Miguel Domingo, Mercedes Garcıa-Martınez, Alexandre Helle, Francisco Casacuberta, and Manuel Herranz. 2018. How much does tokenization affect neural machine translation? *arXiv preprint arXiv:1812.08621*.

Alexander Johannes Edmonds. 2013. The Dialects of Kurdish. *Ruprecht-Karls-Universität Heidelberg*.

Kyumars Sheykh Esmaili, Donya Eliassi, Shahin Salavati, Purya Aliabadi, Asrin Mohammadi, Somayeh Yosefi, and Shownem Hakimi. 2013. Building a test collection for sorani kurdish. In *2013 ACS International Conference on Computer Systems and Applications (AICCSA)*, pages 1–7. IEEE.

Kyumars Sheykh Esmaili and Shahin Salavati. 2013. Sorani Kurdish versus Kurmanji Kurdish: an empirical comparison. In *Proceedings of the 51st Annual Meeting of the Association for Computational Linguistics (Volume 2: Short Papers)*, pages 300–305.

Co-authors Miquel Espla-Gomis, Juan Antonio Pérez-Ortiz, Vıctor M Sánchez-Cartagena, Felipe Sánchez-Martınez, and Reviewers Alexandra Birch. 2019. Global Under-Resourced MEdia Translation (GoURMET). *H2020 Research and Innovation ActionNumber: 825299 - D1.1 – Survey of relevant low-resource languages*.

Mikel L Forcada, Mireia Ginestí-Rosell, Jacob Nordfalk, Jim O'Regan, Sergio Ortiz-Rojas, Juan Antonio Pérez-Ortiz, Felipe Sánchez-Martínez, Gema Ramírez-Sánchez, and Francis M Tyers. 2011. Apertium: a free/open-source platform for rule-based machine translation. *Machine translation*, 25(2):127–144.

Geoffrey Haig and Yaron Matras. 2002. Kurdish linguistics: a brief overview. *STUF-Language Typology and Universals*, 55(1):3–14.

Kanaan M Kaka-Khan. 2018. English to Kurdish Rule-based Machine Translation System. *UHD Journal of Science and Technology*.

Guillaume Klein, Yoon Kim, Yuntian Deng, Jean Senellart, and Alexander M Rush. 2017. Opennmt: Open-source toolkit for neural machine translation. In *Proceedings of ACL 2017, System Demonstrations*, pages 67–72.

Philipp Koehn. 2009. *Statistical machine translation*. Cambridge University Press.

Philipp Koehn. 2020. *Neural machine translation*. Cambridge University Press.

Philipp Koehn and Rebecca Knowles. 2017. Six challenges for neural machine translation. In *Proceedings of the First Workshop on Neural Machine Translation*, pages 28–39.

Alon Lavie and Abhaya Agarwal. 2007. METEOR: An automatic metric for MT evaluation with high levels of correlation with human judgments. In *Proceedings of the second workshop on statistical machine translation*, pages 228–231.

Yinhan Liu, Jiatao Gu, Naman Goyal, Xian Li, Sergey Edunov, Marjan Ghazvininejad, Mike Lewis, and Luke Zettlemoyer. 2020. Multilingual Denoising Pre-training for Neural Machine Translation.

Edward Loper and Steven Bird. 2002. Nltk: the natural language toolkit. *arXiv preprint cs/0205028*.

Minh-Thang Luong, Preslav Nakov, and Min-Yen Kan. 2019. A hybrid morpheme-word representation for machine translation of morphologically rich languages. *arXiv preprint arXiv:1911.08117*.

M. Rita Manzini, Leonardo M. Savoia, and Ludovico Franco. 2015. Ergative case, aspect and person splits: Two case studies. *Acta Linguistica Hungarica Acta Linguistica Hungarica*, 62(3):297 – 351.

Chenggang Mi, Lei Xie, and Yanning Zhang. 2020. Improving adversarial neural machine translation for morphologically rich language. *IEEE Transactions on Emerging Topics in Computational Intelligence*.

Tomas Mikolov, Edouard Grave, Piotr Bojanowski, Christian Puhrsch, and Armand Joulin. 2018. Advances in Pre-Training Distributed Word Representations. In *Proceedings of the International Conference on Language Resources and Evaluation (LREC 2018)*.

George A Miller. 1998. *WordNet: An electronic lexical database*. MIT press.

Preslav Nakov and Jörg Tiedemann. 2012. Combining word-level and character-level models for machine translation between closely-related languages. In *Proceedings of the 50th Annual Meeting of the Association for Computational Linguistics: Short Papers-Volume 2*, pages 301–305. Association for Computational Linguistics.

Kishore Papineni, Salim Roukos, Todd Ward, and Wei-Jing Zhu. 2002. Bleu: a method for automatic evaluation of machine translation. In *Proceedings of the 40th annual meeting on association for computational linguistics*, pages 311–318. Association for Computational Linguistics.

Peyman Passban. 2017. *Machine translation of morphologically rich languages using deep neural networks*. Ph.D. thesis, Dublin City University.

Jeffrey Pennington, Richard Socher, and Christopher D Manning. 2014. Glove: Global vectors for word representation. In *Proceedings of the 2014 conference on empirical methods in natural language processing (EMNLP)*, pages 1532–1543.

Matthew Snover, Bonnie Dorr, Richard Schwartz, Linnea Micciulla, and John Makhoul. 2006. A study of translation edit rate with targeted human annotation. In *Proceedings of association for machine translation in the Americas*, volume 200. Cambridge, MA.

Kaitao Song, Xu Tan, Tao Qin, Jianfeng Lu, and Tie-Yan Liu. 2019. MASS: Masked Sequence to Sequence Pre-training for Language Generation.

Fatima Jalal Taher et al. 2017. Evaluation of inkurdish Machine Translation System. *Journal of University of Human Development*, 3(2):862–868.

Jörg Tiedemann. 2012. Parallel data, tools and interfaces in opus. In *Proceedings of the Eight International Conference on Language Resources and Evaluation (LREC'12)*, Istanbul, Turkey. European Language Resources Association (ELRA).

Sneha Tripathi and Juran Krishna Sarkhel. 2010. Approaches to machine translation. *Annals of Library and Information Studies*, 57:388–393.

Hadi Veisi, Mohammad MohammadAmini, and Hawre Hosseini. 2020. Toward Kurdish language processing: Experiments in collecting and processing the AsoSoft text corpus. *Digital Scholarship in the Humanities*, 35(1):176–193.

Peter W. Smith. 2014. Non-peripheral cliticization and second position in Udi and Sorani Kurdish. In *Paper under revision at Natural Language and Linguistic Theory*, `https://user.uni-frankfurt.de/~psmith/docs/smith_non_peripheral_cliticization.pdf` edition. (Date accessed: 12.05.2020).

Géraldine Walther. 2012. Fitting into morphological structure: accounting for Sorani Kurdish endoclitics. In *Mediterranean Morphology Meetings*, volume 8, pages 299–321. [Online; accessed 19-Mar-2019].

Yonghui Wu, Mike Schuster, Zhifeng Chen, Quoc V Le, Mohammad Norouzi, Wolfgang Macherey, Maxim Krikun, Yuan Cao, Qin Gao, Klaus Macherey, et al. 2016. Google's neural machine translation system: Bridging the gap between human and machine translation. *arXiv preprint arXiv:1609.08144*.

Jiajun Zhang and Chengqing Zong. 2016. Bridging neural machine translation and bilingual dictionaries. *arXiv preprint arXiv:1610.07272*.

# A Appendix

**Tanzil Corpus**

| | |
|---|---|
| en | They are the patient, the sincere and devout, full of charity, who pray for forgiveness in the hours of dawn. |
| ckb | (جا ئەو ئیماندارانە، ئەمە سیفەتیانە) **خۆگرو** ئارامگرن (لە بەرامبەر ناسۆر و ناخۆشیەکانی ژیانەوە)، **ڕاستگۆو** خواپەرستن، ماڵ و سامان دەبەخشن و، لە بەرەبەیانەکاندا داوای لێخۆشبوون دەکەن (لە پەروەردگاریان چونکە واهەست دەکەن کە وەک پێویست خواپەرستیان نەکردووە). |
| ckb norm. | خۆگر و ئارامگرن، ڕاستگۆ و خواپەرستن، ماڵ و سامان دەبەخشن و، لە بەرەبەیانەکاندا داوای لێخۆشبوون دەکەن |
| ckb norm. (WordPunct) | خۆگر و ئارامگرن ، ڕاستگۆ و خواپەرستن ، ماڵ و سامان دەبەخشن و ، لە بەرەبەیانەکاندا داوای لێخۆشبوون دەکەن |
| ckb norm. (Unigram) | خۆگر و ئارامگر ن ، ڕاستگۆ و خواپەرست ن ، ماڵ و سامان دەبەخشن و ، لە بەرەبەیان ەکاندا داوای لێخۆشبوون دەکەن |
| ckb norm. (BPE) | خۆگر و ئارام گرن ، ڕاستگۆ و خوا پەرستن ، ماڵ و سامان دەبەخشن و ، لە بەرەبەیان ەکاندا داوای لێخۆشبوون دەکەن |
| ckb norm. (WordPiece) | خۆگر و ئارامگر ن ، ڕاستگۆ و خوا پەرستن ، ماڵ و سامان دەبەخشن و ، لە بەرەبەیان ەکاندا داوای لێخۆشبوون دەکەن |

**TED Corpus**

| | |
|---|---|
| en | So that instead of spending it the way you usually spend it, maybe if you spent it differently, that might work a little bit better. |
| ckb | کەواتە ئەگەر لە بری ئەوەی بەو شێوازەی ئێستات پارەکە خەرج بکەیت ڕەنگە ئەگەر بە شێوەیەکی تر خەرجی بکەیت ڕەنگە تۆزێک باشتر بێت. |
| ckb norm. | کەواتە ئەگەر لە بری ئەوەی بەو شێوازەی ئێستات پارەکە خەرج بکەیت ڕەنگە ئەگەر بە شێوەیەکی تر خەرجی بکەیت ڕەنگە تۆزێک باشتر بێت . |
| ckb norm. (WordPunct) | کەواتە ئەگەر لە بری ئەوەی بەو شێوازەی ئێستا ت پارەکە خەرج بکەیت ڕ ەنگە ئەگەر بە شێوەیەکی تر خەرجی بکەیت ڕەنگە تۆزێک باشتر بێت . |
| ckb norm. (Unigram) | کەواتە ئەگەر لە بری ئەوەی بەو شێوازەی ئێستا ت پارەکە خەرج بکەیت ڕە نگە ئەگەر بە شێوەیەکی تر خەرجی بکەیت ڕە نگە تۆزێک باشتر بێت . |
| ckb norm. (BPE) | کەواتە ئەگەر لە بری ئەوەی بەو شێوازەی ئێست ات پارەکە خەرج بکەیت ڕ ەنگە ئەگەر بە شێوەیەکی تر خەرجی بکەیت ڕ ەنگە تۆزێک باشتر بێت . |

**KurdNet**

| | |
|---|---|
| en | make less lively, intense, or vigorous; impair in vigor, force, activity, or sensation |
| ckb | کەمتر کردنەوەی وشیاری، زیندووایەتی یان هێز؛ ناکۆکی لە بڕست، هێز، بەکاربوون یان هەستیاری |
| ckb norm. | کەمتر کردنەوەی وشیاری، زیندووایەتی یان هێز؛ ناکۆکی لە بڕست، هێز، بەکاربوون یان هەستیاری |
| ckb norm. (WordPunct) | کەمتر کردنەوەی وشیاری ، زیندووایەتی یان هێز ؛ ناکۆکی لە بڕست ، هێز ، بەکاربوون یان هەستیاری |
| ckb norm. (Unigram) | کەمتر کردنەوەی وشیاری ، زیندوو ایەتی یان هێز ؛ ناکۆکی لە بڕ ست ، هێز ، بەکار بوون یان هەستیاری |
| ckb norm. (BPE) | کەمتر کردنەوەی وشیاری ، زیندوو ایەتی یان هێز ؛ ناکۆکی لە بڕ ست ، هێز ، بەکار بوون یان هەستیاری |
| ckb norm. (WordPiece) | کەمتر کردنەوەی وشیاری ، زیندوو ایەتی یان هێز ؛ ناکۆکی لە بڕ ست ، هێز ، بەکارب وون یان هەستیاری |

Figure A.3: The tokenization of parallel translations of English (en) and Sorani Kurdish (ckb) in the Tanzil, TED and KurdNet–the Kurdish Wordnet. The incorrectly-merged words are indicated in bold and are corrected in the normalized (norm.) step. Tokenization models are specified in parentheses

| | |
|---|---|
| Input (Tanzil) | when they came in to him , and said , salam ! he answered ; salam , and said : you are a people unknown to me . |
| Reference | کاتێک کتوپر خۆیان کرد بە ماڵدا و وتیان : سڵاو ، ئەویش وتی ، سڵاو لەئێوەش بێت ، هەرچەندەناتانناسم . |
| System translation | کاتێک چوون بۆ سەردانی و وتیان : سڵاو ، ئەویش وتی ، سڵاو لەئێوەش بێت ، هەرچەندەناتانناسم . |
| Back-translation | when (they) went to visit him/her and said : hi , then (he) said, hi to you too, although(I)donotknowyou . |
| Input (TED) | all the knowledge and values shared by a society |
| Reference | تەواوی زانست و بەهایانەی کە کۆمەڵ تێدا هاوبەشن |
| System translation | هەر زانیاری و ئامێرێکی تەواو بوون . |
| Back-translation | all the knowledge and a tool of finishing . |
| Input (KurdNet) | a structure consisting of a room or set of rooms comprising a single level of a multilevel building |
| Reference | پێکهاتەیەک کە لە ژوورێک یان چەند ژوور دروست بووە و لە سەر نهۆمێکی بینایێکی چەند نهۆمیە |
| System translation | پێکهاتەیەک کە لە زنجیرەیەک یان لە دیوارێ هاتۆتە دەرێ کە لە ئاستێکی گەورەتردایە |
| Back-translation | a structure that has come out from a chain/a set or a wall that is within a bigger level |

Figure A.4: A few translation examples from English to Sorani Kurdish using the model with the highest BLEU scores (Model 2, WordPunct)

| | |
|---|---|
| Input (Tanzil) | شەرابێکیان پێشکەش دەکرێت ، کە پاک و سەرمۆرە . |
| Reference | they will be given to drink a sealed wine . |
| System translation | they will be given to drink of pure wine , sealed . |
| Back-translation | شەرابێکی خاوێنیان پێشکەش دەکرێت، سەرمۆر . |
| Input (TED) | دەزانیت پێویستت بە چییە ئەنجامی بدەیت چی مومکینە بکرێت ، هەرچەندە سەختیش بێت و هەست دەکەیت کات دیار نامێنێت ، خۆتت بیر دەچێتەوە هەست دەکەیت بەشێکیت لە شتێکی گەورەتر و کاتێ مەرجەکان ئامادە بن ئەو کارەی دەیکەیت خۆی لە خۆیدا دەبێتە شایستەی ئەنجامدان . |
| Reference | you know that what you need to do is possible to do , even though difficult , and sense of time disappears , you forget yourself , you feel part of something larger . and once the conditions are present , what you are doing becomes worth doing for its own sake . |
| System translation | and to show what is said , if you are going to speak , even though you are , even though you are , in fact , a matter is decided , and you will be a leader . |
| Back-translation | و بۆ نیشان دانی ئەوشتەی گوتراوە ، ئەگەر دەتهەوێ قسە بکەی ، تەنانەت ئەگەر قسەش دەکەی ، تەنانەت ئەگەر قسەش دەکەی ، لە ڕاستیدا ، بابەتێک بڕیار دراوە ، و تۆ دەبی بە ڕابەرێک . |
| Input (KurdNet) | نارەزایی دەربڕینی کاریگەران یان کۆمەلانی کەمینە بۆ وە دەست خستنی داخوازیەکانیان |
| Reference | a protest action by labor or minority groups to obtain their demands |
| System translation | the act of expressing a word or phrase or argument for |
| Back-translation | ئەرکی دەبڕینی وشەیەک یان ڕستەیەک یان وتووێژێک |

Figure A.5: A few translation examples from Sorani Kurdish to English using the model with the highest BLEU scores (Model 1, WordPunct)